\documentclass{article}

     \usepackage[preprint]{neurips_2022}

\usepackage[utf8]{inputenc} 
\usepackage[T1]{fontenc}    
\usepackage{hyperref}       
\usepackage{url}            
\usepackage{booktabs}       
\usepackage{amsfonts}       
\usepackage{nicefrac}       
\usepackage{microtype}      
\usepackage{xcolor}         

\usepackage{graphicx}
\usepackage{subcaption}
\usepackage{booktabs} 

\usepackage{algorithm}
\usepackage{algorithmic}

\usepackage{amsmath}
\usepackage{amssymb}
\usepackage{mathtools}
\usepackage{amsthm}

\usepackage{wrapfig}

\title{Temporal Weights}

\author{%
  Adam Kohan \\
  \texttt{akohan@cs.umass.edu} \\
   \And
  Edward A. Rietman\\
  \texttt{erietman@cs.umass.edu} \\
   \And
  Hava T. Siegelmann\\
  \texttt{hava@cs.umass.edu} \\
   \And
   \\
  Biologically Inspired Neural and Dynamical Systems \\
  College of Information and Computer Sciences \\
  University of Massachusetts Amherst \\
  Amherst, MA 01003 \\
}

\begin{document}

\maketitle

\begin{abstract}
In artificial neural networks, weights are a static representation of synapses. However, synapses are not static, they have their own interacting dynamics over time. To instill weights with interacting dynamics, we use a model describing synchronization that is capable of capturing core mechanisms of a range of neural and general biological phenomena over time. An ideal fit for these Temporal Weights (TW) are Neural ODEs, with continuous dynamics and a dependency on time. The resulting recurrent neural networks efficiently model temporal dynamics by computing on the ordering of sequences, and the length and scale of time. By adding temporal weights to a model, we demonstrate better performance, smaller models, and data efficiency on sparse, irregularly sampled time series datasets.
\end{abstract}

\section{Introduction}
\label{introduction}
Time provides an opportunity for neural networks to change their
computation. This change may be in response to the evolving dynamics of
the input distribution, possibly a change in trajectory triggered by
some sparse event, or may be to improve performance by taking another
pass over the same input, but with a different perspective. Static
weights do not adjust for either of these situations and perform the
same computation regardless of the passage of time. A neural network
with static weights has only a form of short term, working memory with
its internal hidden states (nodes) to track time steps, but not
change its computation (weights) on its inputs and memory. In contrast, neural networks with
temporal weights can change how they process their own memory and
inputs over time, not only keep track of time.

At each time step, a neural network with temporal weights changes its
computation by constructing weights $W`$ from a fixed set of parameters $W$,
which is considered to be a form of long term stored memory, and the
current time step $t$: $W^`= S(W,t)$

Here, the weights are a nonlinear function of parameters and time. As a
result, the parameters are shared across time, but the expressivity of
the weight values is not limited to simple linear combinations of
weights or shifts in time. Instead, the weights themselves can capture
some dynamics of the input distribution and form their own trajectory
over time. Given $S$ is non-linear, we are amplifying the effective
amount of weights in the model by apply the same fixed number of parameters differently at each time step. In contrast, static weights are only
the parameters directly applied to the inputs. They can
only capture single values, not any dynamics that are separate from the
layer or network.

There are many options of temporal dynamics for $S$ to model. We use a model of coupled
oscillators describing sychronization behaviors that captures core mechanisms
of a range of neural and general biological phenomena over time. Under this model, weights have an explicit dependence on time and have changing interactions with each other over time. Details are in Section \ref{methods}.

\begin{wrapfigure}{r}{0.55\textwidth}
\begin{center}
\includegraphics[scale=.35,trim = 0 120 378 0,clip]{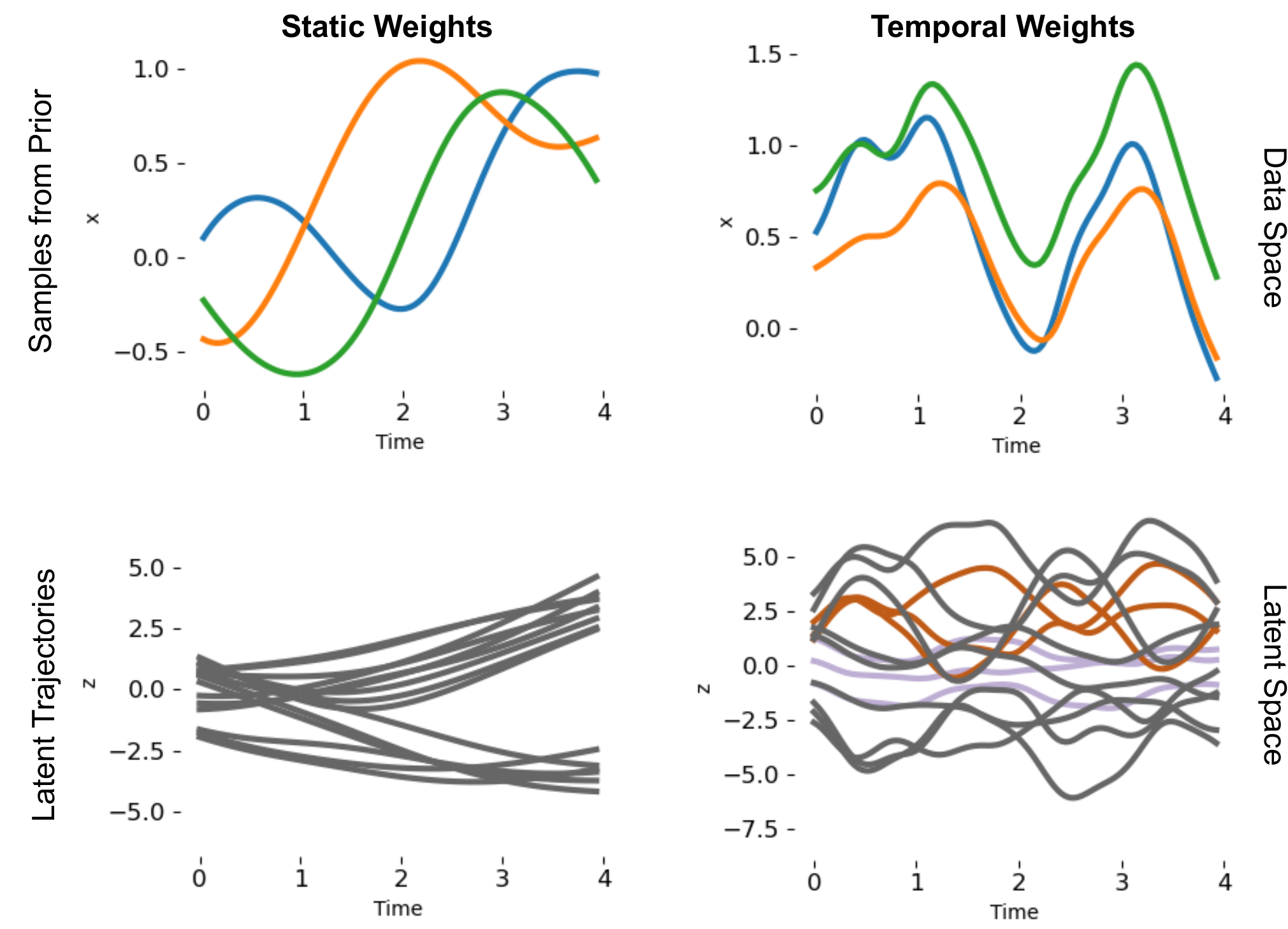}
\caption{A comparison between a neural ode model
with static weights or temporal weights.}
\label{behaviora}
\end{center}
\end{wrapfigure}

To demonstrate the difference between static and temporal weights in practice, we graph in \ref{behaviora} three
samples from a neural ode with static weights (left) and temporal
weights (rights). The neural ode makes for a strong baseline as it
captures temporal trajectories itself and is a state of the art model
for irregularly sampled data. Furthermore, the effect of temporal
weights would have to be non-trivial to alter this trajectory and
significant to improve over it. To ensure the practicality of a
difficult problem, the neural ode models are trained on a sparse,
irregularly sampled dataset (details of the dataset are below). As shown, even a
neural ode model with static weights is not as expressive as the same
model with temporal weights. Notice the well-defined kinks,
quasi-periodicity, and consistent alignment of the temporal weights
model, allowing for a better fit for evolving dynamics and for sparse
and irregularly sampled data. In comparison, the same model with static
weights is inconsistent across samples and has limited smooth
periodicity, which will have more difficulty fitting irregularities or
changes in dynamics over time. We can see that temporal weights
successfully amplifies the parameters to increase the expressivity of
the neural ode model.

\begin{wrapfigure}{l}{0.4\textwidth}
\begin{center}
\includegraphics[scale=.45,trim = 0 147 605 0,clip]{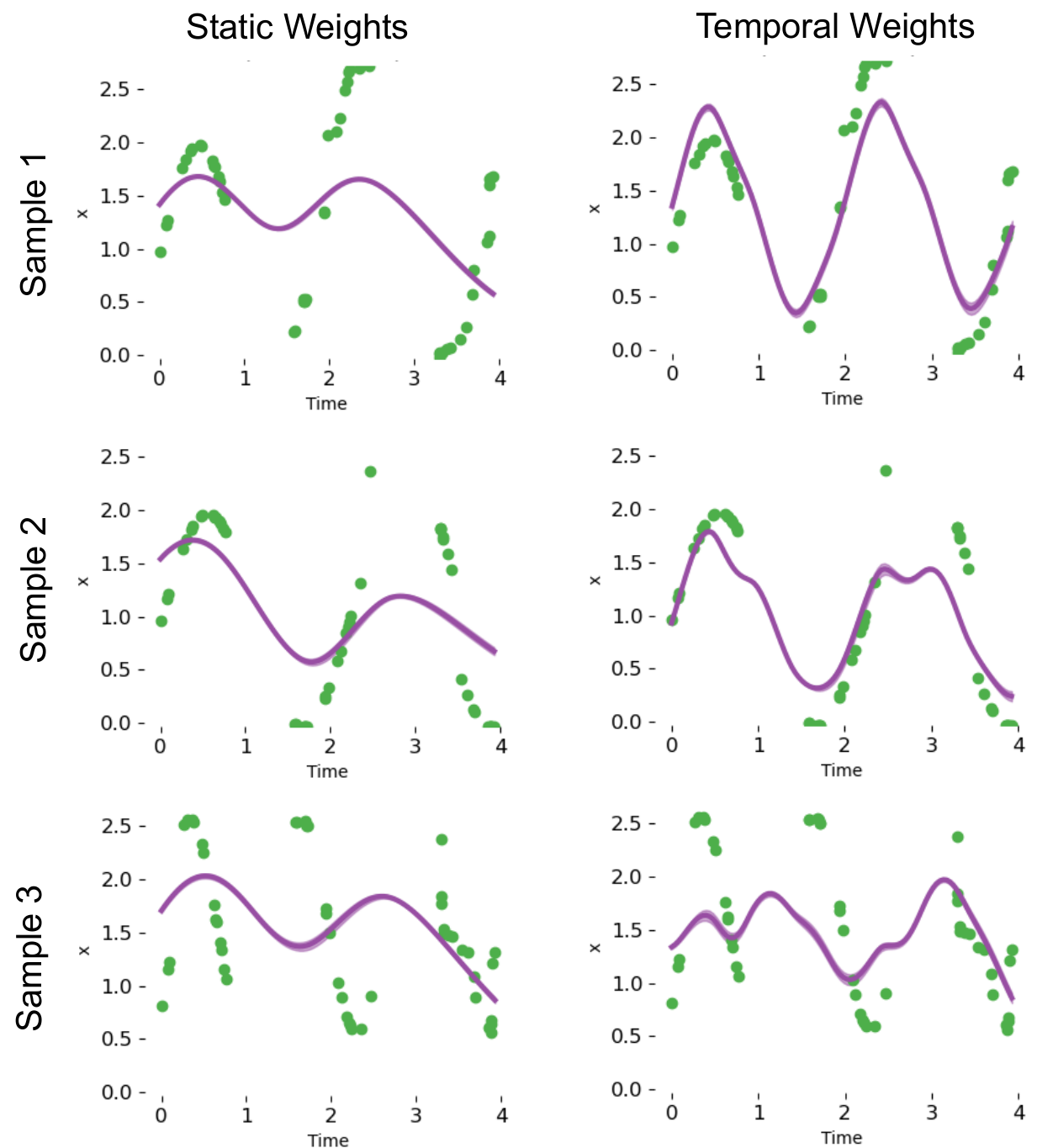}
\caption{Neural ODE model fit on a sparse and
irregular synthetic dataset, with temporal weights or static weights.}
\label{behaviorb}
\end{center}
\end{wrapfigure}

The result of increasing the expressivity of the neural ode model is a
better fit on highly sparse and irregular data, which the original model
struggles to fit. In \ref{behaviorb}, we show results of the neural ode models
on a synthetic dataset of periodic trajectories with variable frequency,
amplitude, noise, and discontinuities. To further increase the
difficulty of this demonstration, the missing points are not used for
training.Fitting a dataset without using the missing points to learn
from is an entirely more challenging task than using them for training
before evaluating the model without them. It more accurately represents
data in the wild, such as the PhysioNet ICU dataset, where the points
are truly missing. So, the model will need to infer the missing points
by referencing other samples for examples of those missing points,
instead of training on them directly. We also limit training to 50
iterations such that overfitting is made difficult and the models must
rely on their efficiency in processing the limited data.

As shown in \ref{behaviorb} (left), the neural ode model with static weights
struggles to fit the data, even though it is able to capture the general
trend. As mentioned above, this model has a limited smooth periodicity.
In contrast, the neural ode with temporal weights, shown in \ref{behaviorb} (right),
is able to efficiently capture the local details of the data. Through
time, this model can continuously perturb its own dynamics and adjust or
change its trajectory. Given the difficulty of this task, neither model
perfectly fits the data. However, the neural ode with temporal weights
clearly performs better, demonstrating more variability across samples
as it adapts to that sample's data points.

\section{Methods}\label{methods}

We developed a neural model whose weights depend explicitly on time and are
. Our
model increases the capacity of the network by incorporating the natural
phenomena of time into the parameters, instead of solely increasing the
number of parameters. In our approach, we use a biological model of time
dependent behavior in neurons as the basis for temporal weights. We
formulate the biological model as a weight scaling algorithm with
oscillatory dynamics for time based reconfigurations of the weights.

\subsection{Neural Synchronization}\label{neural-synchronization}

The scaling function used in our algorithm is based on models of mass
neural synchronization, which has been attributed to play a role in
movement \cite{cassidy2002movement} and memory \cite{klimesch1996memory}.
The dynamic modification of
synaptic connections that is critical to neural synchronization is
recognized to be the basis of long-term memory in learning \cite{abbott2000synaptic,shimizu2000nmda}.
These synaptic dynamics bring about adaptive development of network
structure whose trajectory is oriented by the dependencies of the inputs
on underlying learning mechanisms. Herein, we focus on time-based
dependencies of the underlying learning mechanisms.

\begin{wrapfigure}{r}{0.4\textwidth}
\begin{center}
    \centerline{\includegraphics[scale=.3,trim = 0 352 378 0,clip]{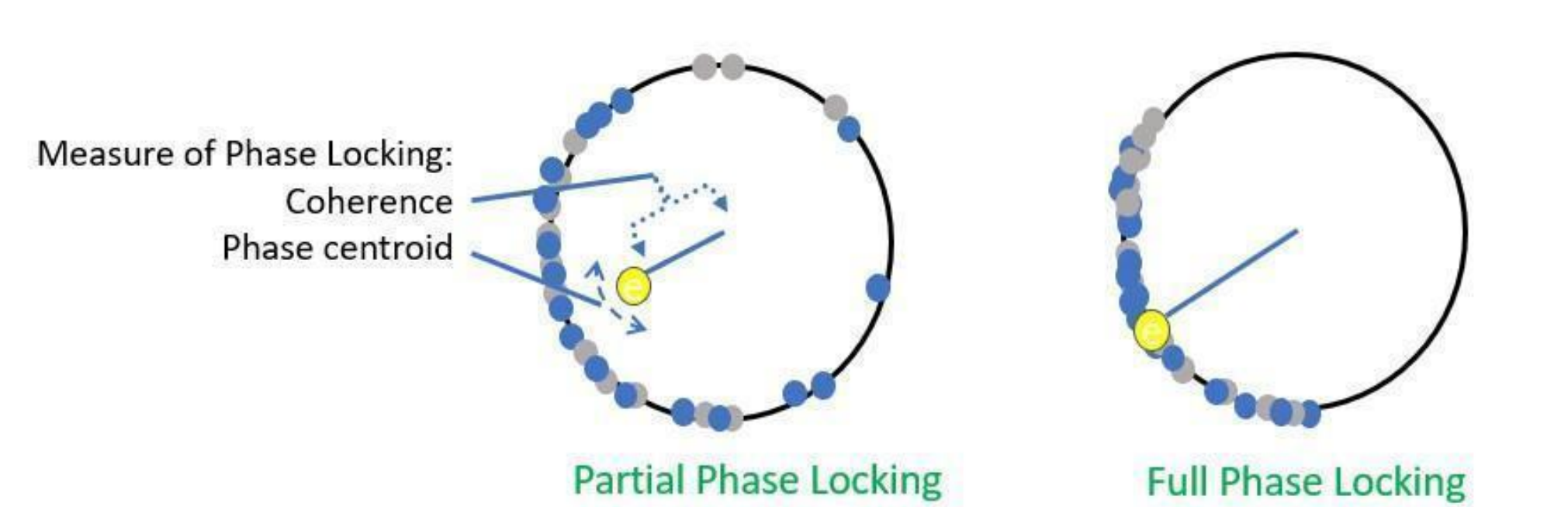}}
\caption{Full and Partial Phase Locking of Weights at Some Time Delta.}
\label{phaselocking}
\end{center}
\end{wrapfigure}

A scaling function $S$ is learned alongside the parameters $W$
of the network. Although $W$ is fixed after training, the network's
temporal weights $W` = S(W,t)$ are different at each time-step due to
the time-dependence of the scaling function. We treat each weight
$w'_{i} \in W`$ as individual units that are controlled by the
underlying mechanisms of mass neural synchronization.
Our scaling function models synchronization behaviors in natural dynamic
systems: it varies the coupling and decoupling of different subsets of
weights given the relative time and the relationship between weights
(Fig. \ref{phaselocking}).

In partial phase locking, clusters of weights are each synchronized to a
different frequency. In full phase locking, most or all of the weights
are synchronized to the same frequency. Our scaling function \(S\) taken
over the weights \emph{W`} learns parameters \(W\) to adjust this phase
locking behavior at each time step. Our scaling function provides a time
based prior for the network. That is, separate weights at each time step
to process each input of a sequence are related to each other and are
non-linear combinations of each other. The network is thus capable of
relating individual weights together and constructing different
combinations of weights to focus on the properties of each input in a
sequence revealed by its point in time. It can isolate different subsets
of weights (i.e. weight configurations) to apply an input-dependent
ordered series of functions to the model's internal state.

\subsection{Neural ODE Model}\label{neural-ode-model}

\begin{figure*}[ht]
\vskip 0.2in
\begin{center}
    \centerline{
        \includegraphics[scale=.3,trim = 0 155 378 10,clip]{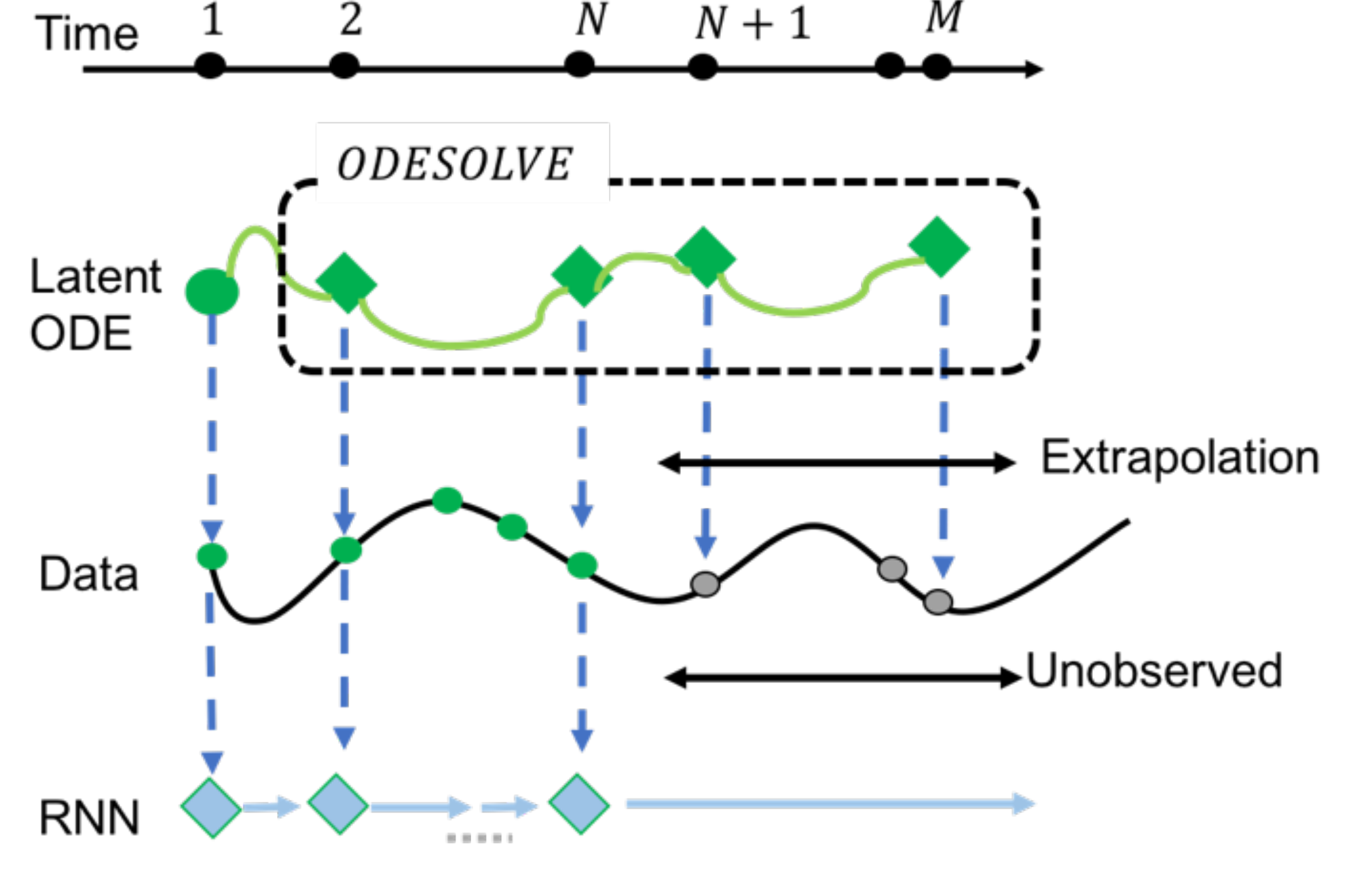}
        \includegraphics[scale=.3,trim = 0 105 315 10,clip]{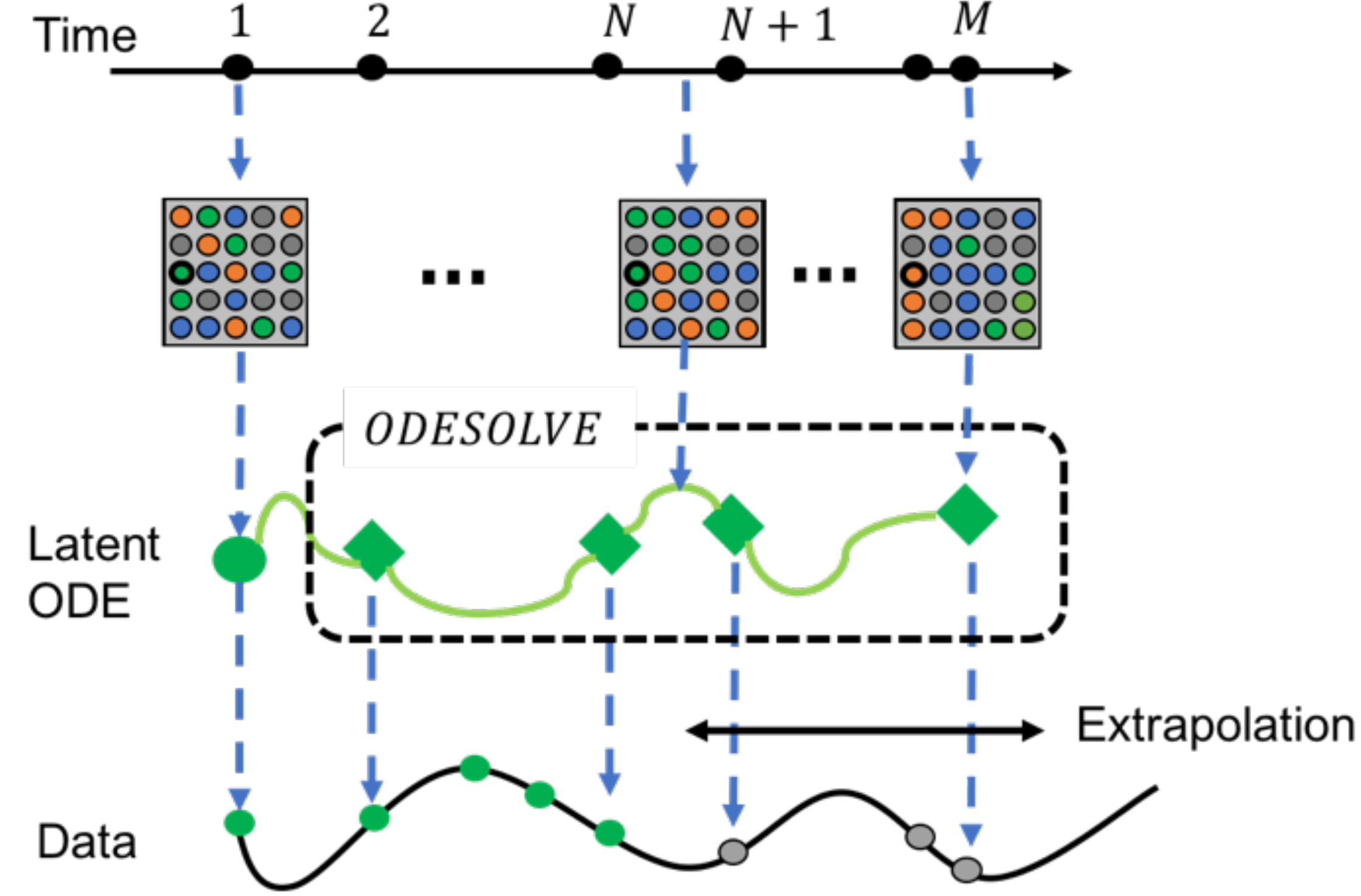}
    }
\caption{Computational Graph of Latent ODE and
Temporal Weights. (Left) The trajectory of the ODE model varies
continuously with time, even at times between receiving inputs. On the
other hand, the state of RNN only changes in response to inputs. (Right)
Temporal Weights construct weights at each time step, which are used
by the Latent ODE model. This includes time steps where there are inputs
or outputs (e.g. time steps $1$, $N$, and $M$), and time steps where there is
neither, but the ode solver is making calls to the neural network, such as
the time step between $N$ and $N + 1$. A view into the ODE solver, denoted by
ODESOLVE in the figures, with temporal weights is shown in Figure 5.}
\label{compgraph}
\end{center}
\vskip -0.2in
\end{figure*}

We apply our scaling function to Neural Ordinary Differential Equation
models \cite{chen2018neural}. Neural ODE models are time continuous models whose
uniquely defined latent trajectory lets us extrapolate predictions
arbitrarily far forwards or backwards in time. In comparison, discrete
models require observations (frequent data points) to control the
trajectory and hence are ill-defined with missing data or irregularity.
Neural ODEs are state of the art for irregularly sampled data \cite{rubanova2019latent}.
Neural ODEs are a natural fit for temporal weights. While Neural ODEs
are time-dependent, their weights are static and only indirectly depend
on time as part of the network's internal state. Our temporal weights
makes weights dependent on time explicitly.

To generate the trajectories between inputs, the Neural ODE model makes
calls to an ODE solver. This solver breaks down the time interval into
smaller subintervals and subsequently approximates the solution to the
ODE at the endpoints of these intervals. Given the solver must call the
scaling function when computing each of these intermediate values, it
also has direct control of the number of weight configurations produced
by from a set of parameters. Depending on the complexity of trajectory
induced by the input, the model will apply a different number of
functions to its internal state. The model can effectively learn when to
increase or decrease the use of its parameters.

\begin{wrapfigure}{r}{0.6\textwidth}
\begin{center}
    \centerline{
        \includegraphics[scale=.3,trim = 0 120 150 0,clip]{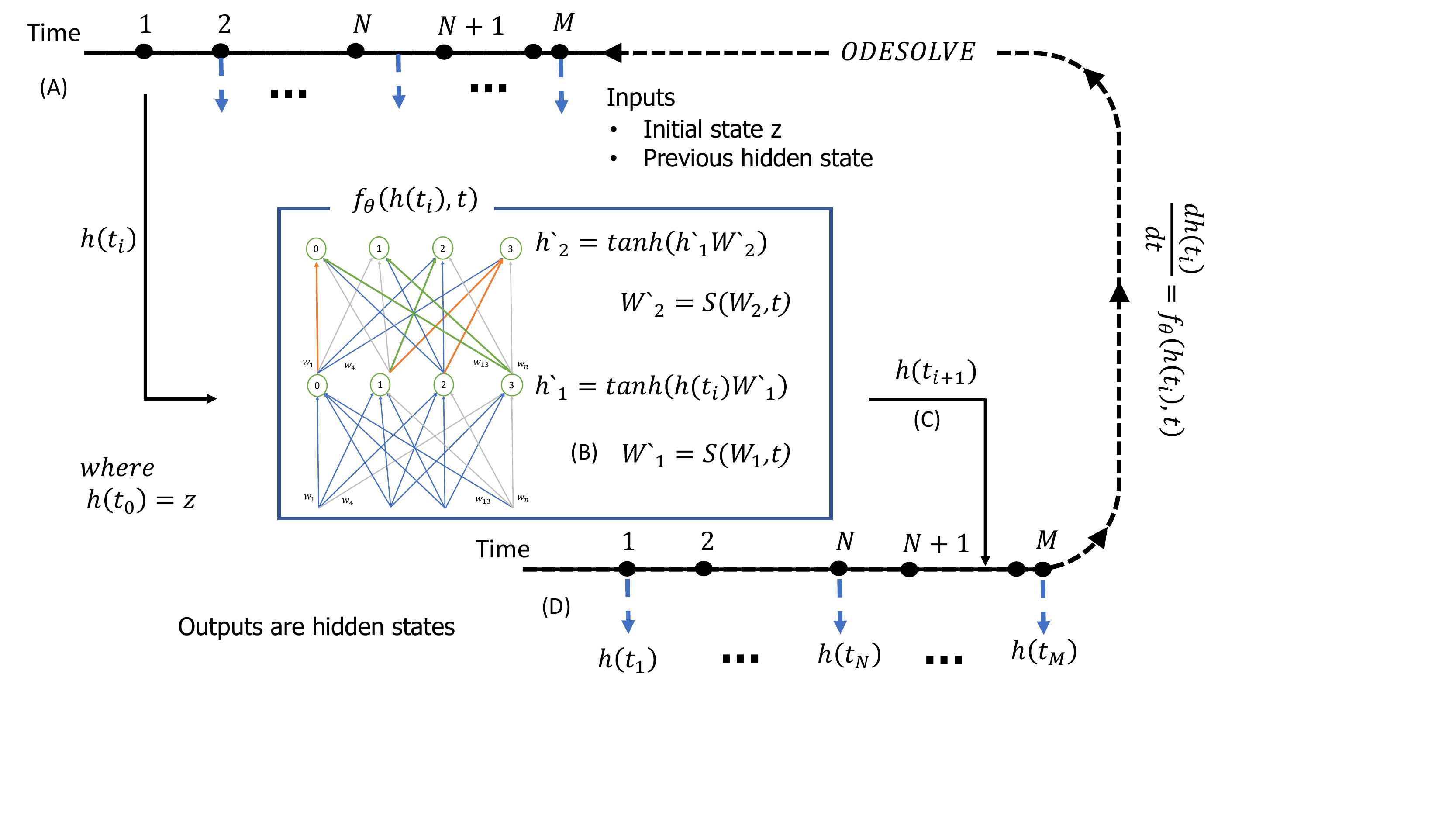}
    }
\caption{Internal Procedure of ODE Solver with Temporal Weights.
The ODE solver calls a multilayer fully connected neural network
internally \(f_{\theta}\). (A) At each input time step, the solver gives
the time \(t_{i}\) and the previous solver's hidden state \(h(t_{i})\)
to the network \(f_{\theta}\). (B) For each layer $l$, the network
constructs weights \({W`}_{l}\) using the parameters \(W_{l}\) and the
time \(t\). The weights \({W`}_{l}\) are used by the layers to process
the network's input \(h(t_{i})\) and the network's previous hidden state
\(h`_{l}\). (C) The network outputs the next hidden state
\(h(t_{i + 1})\), which is given to the ODE solver. (D) If there is an
output time step at \(t_{i + 1}\), \(h(t_{i + 1})\) is also outputted
from the solver. The algorithm is provided in Fig 6.}
\label{odeinternal}
\end{center}
\end{wrapfigure}

The neural ODE is able to control the use of temporal weights through
the ode solver. The ode solver chooses the frequency and the time steps
to call the temporal weight scaling function. So, the neural ODE
automatically adapts the usage temporal weights to the input. In
contrast, discrete neural networks respond directly to observations and
the observations set the frequency and time steps with which to call the
temporal weight scaling function. As a result, the usage of temporal
weights is limited by the number of observations in discrete networks.
To more completely utilize the capacity of temporal weights, we use
neural ODE models.

We apply temporal weights to the Latent ODE and ODE RNN models of \cite{rubanova2019latent}.
Latent ODEs and ODE RNNs are state of the art models on data with arbitrary time gaps
between observations, irregularly sampled data.
The equations and internal procedures of the Neural ODE with temporal
weights used within these models are shown in Figures 4 and 5. Refer to \cite{rubanova2019latent} for further
details of the base models. Unless otherwise stated in \ref{results}, we follow the training procedure and configuration of hyperparameters
from \cite{rubanova2019latent} for consistency. We demonstrate that with temporal
weights, model size can be reduced and have
similar or better performance.

\begin{algorithm}[tb]
   \caption{Neural ODE with Temporal Weights}
   \label{alg:node-tempw}
\begin{algorithmic}
   \STATE {\bfseries Input:} Latent $z$, Output Times $\left\{ t_{i} \right\}_{i = 1\ldots M}$
   \STATE {\bfseries Provide:} Network $f_{\theta}$ with $L$ Layers
   \STATE {\bfseries call} $ODESOLVE(f_{\theta},z,\ \left\{ t_{i} \right\}_{i = 1\ldots M})$

       \begin{ALC@g}
       \FOR{$h\left( t_{i} \right),t$ given by the solver}
       \STATE {\bfseries call} $f_{\theta}\left( h\left( t_{i} \right),t \right)$

           \begin{ALC@g}
           \STATE Let $h`_{0} = \ h\left( t_{i} \right)$

           \FOR{layer $l=1$ {\bfseries to} $L$}
           \STATE $W`_{l} = S\left( W_{l},t \right)$
           \STATE $h`_{l} = \tanh\left( h`_{l - 1}W`_{l} \right)$
           \ENDFOR

           \STATE Output $h\left( t_{i + 1} \right) = h`_{L}$ to the solver
           \end{ALC@g}
       \STATE {\bfseries end call}

       \ENDFOR

       \STATE Return $\left\{ h_{i} \right\}_{i = 1\ldots M}$
       \end{ALC@g}
   \STATE {\bfseries end call}
\end{algorithmic}
\end{algorithm}

\subsection{Temporal Scaling Function}\label{temporal-scaling-function}

The scaling function $S$ drives a sinusoidal wave to scale the
weights. The sine function can be used to output a scalar between
$[0,1]$ or $[-1,1]$ for each weight in the layer. Different weights
will have different magnitudes over time or even be inverted. The output
of the sine function changes with time $t$, given by the ODE
solver. The time is shifting the phase of the sine wave. We let the
network learn parameters to control scale of time for each weight
separately or jointly. That means the network has the capacity to reuse
the same learned dynamics repeating the period over time, or learn to
split up the period over different time intervals, or both. For example,
the network may scale time such that each period \(2\pi\) repeated every
\emph{dt} time interval. Or, the network may split up the period
\(2\pi\) into \(\pi/2\ \) chunks spanning \emph{dt} time intervals. Or,
both. Regardless, the network is able to control the dynamics of the
network through the weights over time. The learned dynamics in the
scaling algorithm are a function of the parameters $W$ and a
comparison of each parameter to all the other weights. The weights
$W`$ for a layer are a function of this comparison, not the actual
weight itself, as shown in Fig 4. The network's weight matrix at each
layer is of interactions between weights, instead of the weights
themselves. The resulting equation is:
\begin{align}
w`_i &= S_{i}\left( W_{l},t \right)\\
&= \sum_{j = 1}^{\# W_{l}}{\frac{K_l^{\left\{ \text{ij} \right\}}}{\# W_{l}}\sin\left\lbrack f_l(t)*\left( w_{i} - w_{j} \right) + \phi_l(t) \right\rbrack}
\end{align}
where \(K_l^{\left\{ \text{ij} \right\}}\) is the parameter matrix of
coupling coefficients, \(\left( w_{i} - w_{j} \right)\) is the
interaction of weight \(i\) with other weights scaled and shifted by the
current time \(t\) in the ODE solver, \(f_(t)\) are the scaling
parameters, and \(\phi_l(t)\) are the shifting parameters. All the
parameters \(K_l^{\left\{ \text{ij} \right\}},\ f_l(t),\ \phi_l(t)\) are
learned with the parameters \(w \in W_l\).

The scaling function takes the time $t$ and parameter matrix
$W_l$ for layer $l$ to construct a weight $w`_i \in W`_l$. The matrix $K$
contains learned coefficients that control the strength of coupling
between weights. The sine function is over the difference between each
weight in a layer. The phase of this function is controlled by $t$.
The result is a periodic estimator of weight distances that evolves as a
function of time $t$.

The scaling function is taken for each layer and each time step as shown
in Fig. 4. Each layer has it's neurons synchronized over time separately
from the other layers. The comparison between all weights in the scaling
function can increase memory usage. Future work may explore a global neural
synchronization across the network if this issue is alleviated.

\section{Experiments}\label{results}

Ten experiments over five datasets with temporally dependent features show
that models with temporal weights more closely capture time dynamics.
They have better performance in fewer epochs and with less parameters
than the same model with static weights. The datasets are sparse and
irregularly sampled to better ensure: (1) the applicability of our model
in the wild where data may be missing, difficult to acquire, or streamed
on-line; and (2) that our model is learning the underlying distributions
over time, not memorizing.

We use the Pytorch library \cite{paszke2019pytorch} and train networks
on a single NVIDIA GeForce GTX TITAN X GPU. We use the Adamax optimizer
with learning rates $0.01$, $0.04$
and exponential learning rate decays $0.999$, $0.9999$. The best
performing model is reported. Individual details are listed
under each experiment. Refer to \ref{neural-ode-model} for additional
information.

\subsection{PhysioNet ICU}\label{physionet-icu}

We evaluate TW on the PhysioNet Challenge 2012 \cite{silva2012predicting}
dataset of 12,000
ICU stays. At admission, a one-time set of general descriptors, such as
age or gender, is collected one. During a stay in the ICU, a sparse
subsets of 37 measurements are taken, such as Lactate, Mg, and NA
levels. Measurements are from the at least the first 48 hours of an ICU
stay. Each stay is labeled whether the patient survives hospitalization
or not. Some measurements are at regular intervals and others are
asynchronous and at irregular times, collected only when required. There
are more than 4,600 possible measurements per time series. To lower
training time, we halved the number measurement times by rounding to the
nearest minute, leaving 2,880 possible measurements per time series. The
dataset is challenging: it is extremely sparse with a missing rate of
around 80\% and has highly imbalanced class distributions with a
prevalence ratio of around 14\%.

\textbf{Mortality Prediction}\label{mortality-prediction}
We look to predict which patients survive hospitalization given the
information collected during the first 48 hours of an ICU stay. We use
Area Under the Curve (AUC) as our performance metric due to the class
imbalance. Results are in \ref{icu-mortality}.
The same model with TW converges to a better performance in fewer epochs
and with less than half the parameters, but is about 1.5x slower per
epoch (refer to \ref{discussion}).

\begin{table}[!htb]

\begin{minipage}{0.45\linewidth}

\caption{PhysioNet ICU}

\begin{subtable}{\linewidth}
\begin{center}
\begin{small}
\begin{sc}
\setlength\tabcolsep{5pt}%
\begin{tabular}{lcccr}
\toprule
Method & AUC & Epochs & \# Params \\
\midrule
L-ODE & 0.857 (0.836) & 80 (31) & 163,972 \\
w/ TW & 0.861 & 31 & 76,427 \\
\bottomrule
\end{tabular}
\end{sc}
\end{small}
\caption{Mortality prediction on the PhysioNet ICU dataset. Given the
first 48 hours of an ICU stay, predict in-hospital mortality.}
\label{icu-mortality}
\end{center}
\end{subtable}

\begin{subtable}[t]{\linewidth}
\begin{center}
\begin{small}
\begin{sc}
\setlength\tabcolsep{5pt}%
\begin{tabular}{lcccr}
\toprule
Method & MSE \tiny{x10\textsuperscript{-3}} & Epochs & \# Params \\
\midrule
L-ODE & 2.280 (2.340) & 59 (49) & 67,071 \\
w/ TW & 1.370 & 49 & 52,016 \\
\midrule
L-ODE & 2.208 (2.300) & 91 (56) & 67,071 \\
w/ TW & 1.900 & 56 & 52,026 \\
\bottomrule
\end{tabular}
\end{sc}
\end{small}
\caption{ICU measurements (Top) Interpolation. (Bottom) Extrapolation. Given the first 24 hours of
measurements, predict the next 24 hours of measurements.
(Parenthesis compare the models at the same epoch.)
}
\label{icu-interp}
\end{center}
\end{subtable}

\end{minipage}%
\hspace{.1\linewidth}
\begin{minipage}{0.45\linewidth}

\caption{MuJoCo Physics Simulation}
\begin{subtable}[t]{\linewidth}
\begin{center}
\begin{small}
\begin{sc}
\setlength\tabcolsep{5pt}%
\begin{tabular}{lcccr}
\toprule
Method & MSE \tiny{x10\textsuperscript{-3}} & Epochs & \# Params \\
\midrule
L-ODE & 3.60 & 94 & 617,619 \\
w/ TW & 3.00 & 91 & 112,739 \\
\bottomrule
\end{tabular}
\end{sc}
\end{small}
\caption{Interpolation of hopper body position. We subsample 10\% of the
time points and predict the other 90\%.}
\label{hopper-interp}
\end{center}
\end{subtable}

\begin{subtable}[t]{\linewidth}
\begin{center}
\begin{small}
\begin{sc}
\setlength\tabcolsep{5pt}%
\begin{tabular}{lcccr}
\toprule
Method & MSE \tiny{x10\textsuperscript{-2}} & Epochs & \# Params \\
\midrule
L-ODE & 1.190 & 93 & 617,619 \\
w/ TW & 1.100 & 99 & 112,739 \\
\midrule
L-ODE & 1.480 & 77 & 617,619 \\
w/ TW & 1.220 & 68 & 112,739 \\
\bottomrule
\end{tabular}
\end{sc}
\end{small}
\caption{Extrapolation of hopper body positions. Given the first half of
the timeline, predict body position in the second half. We subsample
10\% of the time points in the first half of the timeline at each
batch (top) or once when constructing the dataset (bottom).}
\label{hopper-extrap}
\end{center}
\end{subtable}

\end{minipage}

\end{table}

\textbf{Interpolation and Extrapolation}\label{measurement-interpolation-and-extrapolation}
We also look at the task to model the PhysioNet Challenge measurements
data. In this task, we impute missing measurements in each sample to
reconstruct the full set of points in the time series. We report the
mean squared error average over the reconstruction of the dataset.
Results are in \ref{icu-interp}.
In the extrapolation task, we provide the network with the first half of
the timeline and construct measurements in the second half of the
timeline. Results are in \ref{icu-interp}.
The latent ODE model with temporal weights converges to a significantly better performance in both the extrapolation and interpolation tasks.

\textbf{Architecture}
The Latent ODE is as follows:
an encoder comprised of a GRU of $50$ units, hidden size of $40$, and Neural ODE of size $50$ with $3$ layers;
a latent size of $20$; and a decoder comprised of a Neural ODE of size $50$ with $3$ layers.
The Latent ODE with temporal weights is as follows:
an encoder comprised of a GRU of $25$ units, hidden size of $20$, and Neural ODE of size $25$ with $3$ layers;
a latent size of $20$; and a decoder comprised of a Neural ODE of size $25$ with $3$ layers.

\subsection{Human Activity}\label{human-activity}

We evaluate TW on the Human Activity dataset \cite{kaluvza2010agent} of
5 individuals performing
7 activities, such as sitting and walking, resulting in 25 sequences of
6,600 time points each on average. 6,554 sequences of 211 time points
each. The dataset has 12 features of 3D positions from tags attached to
a belt, chest, and each ankle (3 dimensions x 4 tags).

\textbf{Time-Point Classification}\label{time-point-classification}
The task is to classify each time point into one of the activities for a
total of 1,477 (7 x 211) outputs.
The model with temporal weights has better performance, converges in
fewer epochs. Results are in \ref{humact-act}.

\begin{table}[!htb]

\begin{minipage}{0.45\linewidth}
\begin{center}
\caption{Human Activity. Classification at each time point.
(Parenthesis compare the models at the same epoch.)}
\label{humact-act}
\begin{small}
\begin{sc}
\setlength\tabcolsep{4pt}%
\begin{tabular}{lcccr}
\toprule
Method & Acc & Epochs & \# Params \\
\midrule
L-ODE & 0.846 (0.748) & 52 (10) & 1,696,763 \\
w/ TW & 0.870 & 10 & 141,023 \\
\bottomrule
\end{tabular}
\end{sc}
\end{small}
\end{center}
\end{minipage}%
\hspace{.07\linewidth}
\begin{minipage}{0.45\linewidth}
\begin{center}
\caption{Climate prediction.}
\label{ushcn-pred}
\begin{small}
\begin{sc}
\setlength\tabcolsep{4pt}%
\begin{tabular}{lcccr}
\toprule
Method & MSE & NLL & Epochs & \# Params \\
\midrule
GRUODE & 0.43 & 0.84 & 74 & 42,640 \\
w/ TW & 0.40 & 0.87 & 68 & 9,105 \\
\bottomrule
\end{tabular}
\end{sc}
\end{small}
\end{center}
\end{minipage}

\begin{minipage}{0.45\linewidth}
\begin{center}
\caption{Sepsis prediction. At every hour, predict whether the patient
will have sepsis within the next 6 to 12 hours. (Parenthesis compare the models at the same epoch.)}
\label{sepsis-pred}
\begin{small}
\begin{sc}
\setlength\tabcolsep{4pt}%
\begin{tabular}{lcccr}
\toprule
Method & AUC & Epochs & \# Params \\
\midrule
ODERNN & 0.689 (0.669) & 67 (24) & 148,672 \\
ODERNN & 0.765 (0.696) & 86 (24) & 680,462  \\
w/ TW & 0.787 & 24 & 149,519 \\
\midrule
NCDE & 0.925 (0.905) & 130 (110) & 55,949 \\
NCDE & 0.925 & 110 & 193,541  \\
w/ TW & 0.931 (0.918) & 180 (110) & 58,553 \\
\bottomrule
\end{tabular}
\end{sc}
\end{small}
\end{center}
\end{minipage}

\end{table}

\textbf{Architecture}
The Latent ODE is as follows:
an encoder comprised of a GRU of $50$ units, hidden size of $100$, and Neural ODE of size $500$ with $4$ layers;
a latent size of $15$; and a decoder comprised of a Neural ODE of size $500$ with $2$ layers.
The Latent ODE with temporal weights is as follows:
an encoder comprised of a GRU of $25$ units, hidden size of $20$, and Neural ODE of size $25$ with $3$ layers;
a latent size of $20$; and a decoder comprised of a Neural ODE of size $25$ with $3$ layers.

\subsection{MuJoCo Physics Simulation}\label{mujoco-physics-simulation}

The physics simulation dataset \cite{rubanova2019latent} is 10,000 sequences of 100 regularly
sampled 14-dimensional time-points generated from the MuJoCo \cite{todorov2012mujoco} simulator.
The dataset is of a bipedal model, called the Hopper model in the
Deepmind Control Suite \cite{tassa2018deepmind}. The task is to learn to approximate Newtonian
physics of the hopper rotating in the air and falling to the ground
starting from a randomly sampled position and velocity. The resulting
deterministic trajectories are dictated by their initial states. This
should be straightforward for the Latent-ODE model to fit as it matches
the assumptions made by the Latent-ODE model, it is dictated by its
initial conditions.

In the \textbf{interpolation task}, we subsample 10\% of the time points to
simulate sparse observation times, and predict the other 90\% time
points. Results are in \ref{hopper-interp}.
In the \textbf{extrapolation task}, we provide the network with the first half of
the timeline, and then predict the second half. We subsampling 10\% of the
time points from the first half of the timeline, but predict on all the points
in the second half of the timeline. We show results in for \ref{hopper-extrap}
two versions of this task. In the first version, shown in the top rows of the table,
we subsample 10\% of the time points at every batch. This means that the networks
will through multiple epochs eventually see all the data points of an input since
a different 10\% of the time points are sampled each epoch. In the second version,
shown in the bottom rows of the table, we subsample 10\% of the time points once when
the dataset is constructed such that the network will never see the missing the
time points. The second version of this task is a more difficult as the missing
time points are never used for training. We see in the results that the performance
gap between the model with and without temporal weights doubles.
In these tasks, we demonstrate that the latent ODE models performs
better with temporal weights, and has 5 times fewer parameters. Our
model has a similar capacity to the larger model, but with much fewer
parameters.

\textbf{Architecture}
The Latent ODE is as follows:
an encoder comprised of a GRU of $100$ units, hidden size of $30$, and Neural ODE of size $300$ with $3$ layers;
a latent size of $15$; and a decoder comprised of a Neural ODE of size $300$ with $3$ layers.
The Latent ODE with temporal weights is as follows:
an encoder comprised of a GRU of $50$ units, hidden size of $40$, and Neural ODE of size $50$ with $3$ layers;
a latent size of $15$; and a decoder comprised of a Neural ODE of size $50$ with $3$ layers.

\subsection{PhysioNet Sepsis}\label{physionet-sepsis}

The PhysioNet Sepsis dataset contains time series of observations from 40,000
ICU patients that were aggregated from two different U.S. hospitals. Each
patient has demographics as a general descriptor, such as agent and gender.
During a stay in the ICU, up to 40 measurements are taken, such as vital signs
and laboratory results. Measurements were recorded together every hour. Each
hour is labeled whether or not an onset of sepsis occurred. The dataset
is sparse with a missing rate of around 74\% and has imbalanced
class distributions with a prevalence ratio of around 7.3\%.

\textbf{Sepsis Prediction}\label{sepsis-prediction}
The task is to predict at every hour whether the patient will have sepsis
within the next 6 to 12 hours. Results are in \ref{sepsis-pred}.
The model with temporal weights has better performance with a similar number of parameters.

\textbf{Architecture}
The smaller ODE RNN with and without TW is as follows:
an encoder comprised of a GRU of $50$ units; a latent size of $32$;
a decoder comprised of an MLP of $100$ units with $2$ layers; and a Neural ODE of size $10$ with $1$ layer.
The larger ODE RNN without TW is as follows:
an encoder comprised of a GRU of $1024$ units; a latent size of $32$;
a decoder comprised of an MLP of $100$ units with $2$ layers; and a Neural ODE of size $20$ with $1$ layer.

\subsection{USHCN Daily Climate Data}\label{ushcn-climate}
The United State Historical Climatology Network (USHCN) daily dataset \cite{menne2010long} contains measurements of 5 climate variables (temperatures, precipitation, and snow) over 150 years for 1,218 meteorological stations. We use the processed data from \cite{de2019gru}. It contains a subset of 1,114 stations over an observation window of 4 years subsampled such that each station has around 60 observations on average. The task is to predict the next 3 measurements after the first 3 years of observation.

\textbf{Architecture}
The smaller GRUODE with TW is as follows:
A p\_model with $2$ layers of $25$ units and an input of $15$ units.
A classification network with $2$ layers of $2$ units and an input of $50$ units.
A gru with $3$ layers of $15$ units and $2$ layers of $45$ units.
A covariates map with $2$ layers of $50$ units and an output of $15$ units.

The larger GRUODE without TW is as follows:
A p\_model with $2$ layers of $25$ units and an input of $50$ units.
A classification network with $2$ layers of $2$ units and an input of $50$ units.
A gru with $3$ layers of $50$ units and $2$ layers of $150$ units.
A covariates map with $2$ layers of $50$ units and an output of $50$ units.

\section{Related Works}\label{related}

Temporal weights are a model of dendrites and synapses. The common view is that
memories are encoded in the connection strength between neurons. In artificial
neural networks, static weights represent (are inspired by) synapses.
However synapses are not static, they have their own dynamics, in
addition to the dynamics of the neurons they connect to.

In \cite{poirazi2001impact},
the authors investigate the storage capacity of neurons that have synapses
with their own dynamics. The authors compare (1) synapses that are linearly summed
across dendritic arbors (contacts with neurons) with (2) synapses where the dendritic
compartments are non-linear functions. The results demonstrate a much larger capacity
for neurons with non-linear synaptic and dendritic subunits. This suggests that
long term memory storage is an active process encoded in the dynamics of synapses and
dendrites. Our results support these findings: a neural network with temporal weights
has similar (or better) performance to a larger version of same network without
temporal weights.

Given the increased capacity from the non-linearity, in temporal weights, we makes this
capacity accessible to the neural network for processing inputs by modeling the behavior
of synaptic plasticity over time. Synaptic plasticity is commonly known for mediating
learning and behavior by changing synaptic strength given a local or global signal or an input.
This mediation can be viewed as a form of selective content addressing. We take a look at
synaptic plasticity from this view, as a mechanism to make information available to the network.
Specifically, we take this view while looking at the temporal behavior of synaptic plasticity
in changing synaptic strength. Synaptic plasticity and in turn changes in synaptic strength are
usually looked at in direct response to some stimuli. However, we look at
similar response patterns to stimuli over time.

In \cite{dobrunz1999response}, the authors stimulated synapses with natural
patterns derived from in vivo recordings. These natural patterns do not have constant frequency.
They provide a better dataset to judge the relative importance of short- (and long-) term
synaptic plasticity for usual synaptic function. The authors find that synaptic strength is
being modulated by the timing of the stimulus. Similar changes in synaptic strength occur
at different time scales given input history. In temporal weights, we capture this
behavior in two approaches. First, the scaling function is based on models of
mass neural synchronization whose behavior oscillates over time. Second, we include
additional parameters in the scaling function for the network to learn a scaling and
length of time for the dataset. In total, the network is able to change its weights
based on the timing of an observation in the input sequence.

Temporal weights work similarly to synaptic plasticity mechanisms. Synaptic plasticity
mechanisms have been applied to neural networks for memory dependent tasks such as catastrophic
forgetting and more recently continual and lifelong learning \cite{miconi2018differentiable,ba2016using}.
In general, these approaches use synaptic plasticity to help the network retain and consolidate memories from
previous data while training on new data, though the goal for this process varies amongst tasks. In contrast to
these works, temporal weights help the network learn the scaling and length of time underlying the trends
in the dataset, which is an active pattern shared across information, not the information itself. Though,
there are functional similarities. In these approaches and temporal weights, the networks learns how accomplish
their respective tasks using parameters, instead of specifying a fixed behavior.
Temporal weights has a similar formulation to \cite{davis2020time,ha2016hypernetworks}.
In \cite{davis2020time} weight are explicitly dependent on time where the weights are linear functions of time,
however we model weights as synapses and dendrites, and are non-linear functions.
In \cite{ha2016hypernetworks}, the weights are non-linear functions as the weights are generated by additional networks
using the previous hidden state as input, and so are indirectly dependendent on time.

\section{Discussion}\label{discussion}

We begin by motivating discussion on the substance of temporal data, namely by distinguishing it from other forms of data. Temporal data is different from sequential data or sequential data paired with time stamps (pseudo-temporal data). Sequential data includes tasks such as copy, associative recall, sorting, frequency, n-mnist, moving-mnist, and split-mnist that were made popular by the neural memory, neural attention, neural ODE, and continual learning literatures (cite neural turing machine work). Here are two examples. Under memory tasks, the moving-mnist is generated from moving digits across a frame. Under attention related tasks, the n-mnist dataset is generated from a sensor detecting a change in an image. With these descriptions of the tasks, it is clear that the substance of these data is memory and attention. Furthermore, it is intented as such in their relevant works. Yet, multiple works we cited use sequential data as a stand-in for temporal data. In contrast, temporal data is generated from dynamic systems, that is a time dependence. Sequential data does not have a time dependence. (Temporal causality.)

However, there is overlap when distinguishing between sequential and temporal data, and sequential data can resemble temporal data. Sequential data and temporal data start to overlap when considering event based data which resembles both temporal data and sequential data. We can allievate some of the overlap by conditioning on attention based data, such as the n-mnist dataset. That is, event based data that is generated from changes in attention on an underlying process is sequential data if the underlying process is not a dynamical system. In other words, it depends on what we are paying attention to. This recursive definition is intentional as it makes clear that event and attention based data are themselves either temporal or sequential data.

Another and very similar point of overlap is in trajectory data. Well recognized properties of dynamical systems are quasi-periodicity, almost periodicity, and regular periodicity. However, periodicity does not imply a dynamical system. This is an important consideration for trajectory data where frequency tasks, such as those using wave or spiral datasets, resemble temporal dyamical processes, but lack the substance of dynamical systems. For example, learning to contruct periodic trajectories or distinguishing between periodic trajectories are memory and attention tasks, respectively. They do not have a time dependence and are not capturing dynamical systems.

We encourage the community to consider the following when naming data as temporal: (1) are underlying dynamical systems generating the data, (2) has the temporal dimensionality of the data been reduced such that it is no longer part of the dataset, (3) is the task online or offline, (4) if it online, at what scale and duration is the input before making the prediction and the prediction itself. These points distinguish temporal data from sequential data and provide context on the nature of time in the temporal dataset. Of greater interest and importance, a direct acknowledgment of time bring us into lifelong learning where consideration of time is essential for negotiating through a deep sea of dynamical systems, where every entity is online except when it is protected.

Our temporal weights significantly increases the expressiveness of the
Neural ODE model by learning to dynamically reconfigure the weights
over time using a model of neural synchronization. We demonstrate that the
resulting models have better performance, fewer parameters, and improved
data efficiency. The Neural ODE model with temporal weights performs much better
on generative prediction tasks, such as the PhysioNet ICU data
imputation task. The neural synchronization behavior seems be to able to
better fit the underlying distribution of the data
as it can learn the temporal patterns in the data and change the
weights accordingly at each time step.

In the Hopper and PhysioNet experiments, we demonstrate that temporal
weights has a capacity better than simply increasing the number of
parameters. The model of neural synchronization better captures the
Newtonian physics of the hopper rotating in the air and falling to the
ground, and the underlying temporal distribution of the ICU measurements.

In the Human Activity and PhysioNet experiments, models with temporal weights have better or
similar performance than the state of the art models, but in fewer
epochs and less parameters. That is, there is better usage of the data during learning.
However, the addition of nonlinear weights increases the time of each
training epoch by $1.2-2.0$ times. This is partially due to the model of mass
neural synchronization requiring comparisons between all the weights, and then the
backward pass needing to compute the gradients for each of these
comparisons. The temporal weights may also be straining the ODE solver
since the weights are no longer fixed every time the solver calls the
network. In general, Neural ODE models are slow to train due to the solver and it is a
trade-off for using these models. Discrete networks do not have this issue.
We are working on expanding temporal weights to discrete networks.

By introducing time as an explicit dependency of the weights, we have
demonstrated that Neural ODE models can better capture the temporal
dynamics of a dataset. We have chosen smaller
model sizes and datasets
with sparse, irregularly sampled time points to elucidate that model with temporal weights
are not memorizing, but have an improved capacity for learning.

{\small
\bibliographystyle{unsrtnat}
\bibliography{./arxiv}
}

\appendix
\newpage

\section{Appendix}

We provide comparisons to additional models in the tables below.

\begin{table}[!htbp]

\begin{minipage}{0.45\linewidth}
\begin{center}
\caption{Classification PhysioNet ICU.}
\label{icu-mortality-perf}
\begin{small}
\begin{sc}
\setlength\tabcolsep{4pt}%
\begin{tabular}{lcccr}
\toprule
Method & AUC \\
\midrule
Latent ODE & 0.857 \\
w/ TW & 0.861 \\
\midrule
SAITS & 0.848 \\
BRITS & 0.835 \\
GP-VAE & 0.834 \\
M-RNN & 0.822 \\
E2GAN & 0.830 \\
GRUI-GAN & 0.830 \\
GRU-D & 0.863 \\
GRU-Simple & 0.808 \\
IP-NETS & 0.860 \\
PHASED-LSTM & 0.790 \\
SEFT-ATTN & 0.851 \\
Transformer & 0.863 \\
\bottomrule
\end{tabular}

from \cite{horn2020set,du2022saits}
\end{sc}
\end{small}
\end{center}
\end{minipage}%
\hspace{.07\linewidth}
\begin{minipage}{0.45\linewidth}
\begin{center}
\caption{Classification on PhysioNet Sepsis}
\label{icu-sepsis-perf}
\begin{small}
\begin{sc}
\setlength\tabcolsep{4pt}%
\begin{tabular}{lcccr}
\toprule
Method & AUC \\
\midrule
ODERNN & 0.765  \\
w/ TW & 0.787 \\
\midrule
NCDE & 0.925 \\
NCDE & 0.925  \\
w/ TW & 0.931 \\
\midrule
GRU-D & 0.674 \\
GRU-SIMPLE & 0.781 \\
IP-NETS & 0.742 \\
PHASED-LSTM & 0.754 \\
SELF-ATTN & 0.768 \\
Transformer & 0.658 \\
\bottomrule
\end{tabular}

from \cite{horn2020set}
\end{sc}
\end{small}
\end{center}
\end{minipage}

\begin{minipage}{0.45\linewidth}
\begin{center}
\caption{Number of Parameters (Rounded)}
\label{icu-impute-num-params}
\begin{small}
\begin{sc}
\setlength\tabcolsep{4pt}%
\begin{tabular}{lcccr}
\toprule
Method & \#Params \\
\midrule
L-ODE & 67,071 \\
w/ TW & 52,016 \\
\midrule
M-RNN & 70,000\\
E2GAN & 80,000\\
GP-VAE & 150,000\\
GRUI-GAN & 160,000\\
BRITS & 730,000\\
Transformer & 4,360,000\\
SAITS & 5,320,000\\
\bottomrule
\end{tabular}
\end{sc}
\end{small}
\end{center}
\end{minipage}

\end{table}

\end{document}